# Exploiting Uncertain and Temporal Information in Correlation

**J.Bigham**
Department of Electronic Engineering,
Queen Mary & Westfield College, London University,
Mile End Road, London E1 4NS, U.K.

## Abstract

A modelling language is decribed which is suitable for the correlation when the underlying functional model of the system is incomplete or uncertain and the temporal dependencies are imprecise. An efficient and incremental implementation is outlined which depends on cost functions satisfying certain criteria. Possibilistic logic and probability theory (as it is used in the applications targetted) satisfy these criteria.

## 1 INTRODUCTION

A modelling language is decribed which is suitable for the correlation of information when the underlying functional model of the system is incomplete or uncertain and the temporal dependencies are imprecise. This language has its roots in the diagnosis and maintenance of telecommunications systems and is designed to be modular and to allow the modelling of large systems. The need to integrate uncertainty management and temporal constraint management is illustrated in the context of diagnosis in a satellite subsystem. An efficient implementation approach is outlined using cost functions which satisfy certain criteria to control computation. Possibilistic logic and probability theory (as it is used here) do.

Models are initially constructed in a logical form in terms of covering rules. Logical relationships can be used to represent several different uncertainty calculi, including probability, belief functions and possibilistic logic, and to model the temporal constraints. The result of a diagnosis is the generation of explanations that explain all the symptoms, at least to some degree of belief. An explanation is a conjunction of necessary contributory causes each with a temporal imprecision constraint and a degree of belief. Multiple causes are allowed.

### 1.1 MOTIVATION

In many on-line diagnostic systems fault reports are collected and clustered according to the time in which they occur. An assumption of such systems is that there are no significantly differing propagation delays along different causal paths. However in some applications there may be imprecise or indeterminate time delays along different causal paths. In such cases the clustering and correlation processes have to be integrated.

*Example: Different time delays in a thermal and electrical causal path*

Consider an example as shown in figure 1 that sketches a fault impact model some of the equipment connected to the power bus of a satellite. The uppermost shaded nodes represent the possible causes. Causal relationships between events are qualified with uncertainty degrees (here possibility $\Pi$ and necessity N) and temporal delays are represented as time intervals [d-min, d-max], representing minimum and maximum delays between consecutive events. The uncertainty in the delay rules are interpreted as the belief in the transition. The interval effectively gives the time period during which the transition must happen if it is going to happen. For example after a temperature increase in the KU area begins (at time t say) it can take between 5 and 60 minutes to cause watchdog software triggering. The necessity of the triggering is $\geq 0.9$, but if it is to trigger it must happen between [t+5, t+60].

A Voltage Regulator connects enough Solar Arrays Sections to the power bus to meet the power consumption, and batteries are used to provide shortfall in the solar energy. The causal model expresses the knowledge about uncertain links between faults and observable events - here using possibilistic logic, and the imprecise temporal delays between the events. The model describes the dynamic behaviour of the power subsystem of the satellite. For example the failure of the protection transistor may have both an electrical and thermal impact on the satellite, with different dynamics. Electrical propagation is faster than thermal propagation. More precisely, if the Over-Voltage transistor fails then the symptom "payload shedding" can arise via two different causal paths. In the illustration the joint possibility distributions are essentially 'noisy or' logic, though any distribution can be modelled. Exploiting uncertainty in the causal links allows a finer description of the phenomena and the ranking of the fault hypotheses.

Shedding occurs when triggered. Only the first signal to shed payload can be observed. So the signal to shed as a result of thermal effects can be masked by the previous electrical signal. However the electrical path is not certain so sometimes the first signal is caused by the thermal effects. If we do not account for the time delays, then observing payload shedding may lead us to give too much belief to Over-Voltage Transistor failure, rather than the other possible causes -- an *exceptional payload power request* or *a problem in the KU area*. Using the necessities (or probabilities in another modeling) ignoring the temporal constraints is basing the diagnosis on the assumption that every possible symptom that could be observed had enough time to be observed.

Identifying possible causes possible causes early, allows us to predict possible future events and look for them, and take corrective action. The temporal information gives extra discrimination power.

As mentioned above the above fault impact model has two ways that payload shedding can occur. Certainly to compute the belief in payload shedding this loop is important as the probability of the disjunction needs to be calculated. However when the objective is to compute possible explanations for payload shedding then since the different possible paths to payload shedding correspond to different causal paths, it is appropriate for an explanation to be a conjunction of necessary conditions along a single *causal* path. The efficiency of the algorithm described to compute explanations depends on this assumption in the case of probability theory, though for possibilistic logic it does not matter.

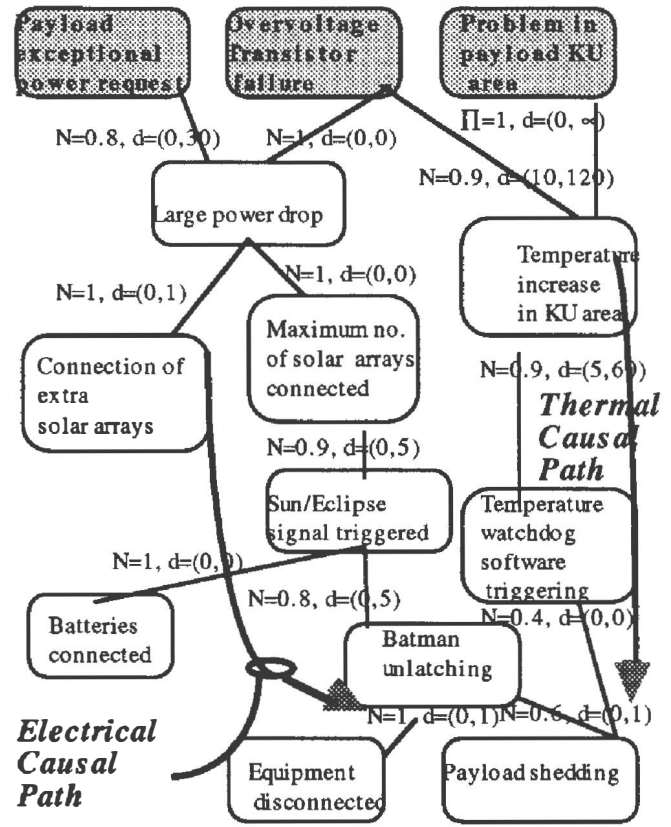

Figure 1: A simplified fault impact model including uncertain and temporal information

## 2 THE MODELING LANGUAGE

A language suitable for representing behaviour relevant to diagnostic reasoning when that reasoning uses logical relationships, uncertain dependencies, and temporal delays, is described.

A functional entity consists of *input ports* , *output ports* and internal attributes called *states*. State variables and ports have a user defined set of mutually exclusive values such as {working, not-working} or {high, medium, low}. The ports are connected to the ports of other functional entities, e.g. the power-out port of a modelled converter is connected with the power-in port of a modelled multiplexor-group. The functional entities together with these port-to-port connections are the functional model. The functional entities are called *units* in the modelling language. The state variable usually represents some unobservable factor which determines the behaviour. Sometimes a form of state variable is used to represent an environmental condition, such as a particular configuration or setting of the system. These are then called environmental variables in the implementation. Ports are labelled locally within a functional entity and so the full identification of a port is something like F.O, where F is the name of the functional entity and O is the name of a port of the functional entity F. There may be other ports called O in different functional entities. State variables have a similar naming scheme.

When building functional models from units an output port may be linked to any number of downstream input ports, but any input port may only be connected to one upstream output port. This is to ensure that all behaviour is expressed in the units themselves and not in the connections between the ports. The connections between ports simply indicates a correspondence, and it is this separation which allows behaviour to be described locally. Units can be held in libraries which can be re-used at different places in the model, e.g. an optical link unit may be used whenever that kind of optical link is required. An output port and its associated input port(s) must all have the same domain. The representation has been used to support fault diagnosis in telecommunications systems (Bigham).

Extending the notation of Konolige to temporal and uncertain models, the functionality of a diagnostic tool adopting a causal model of the unit port type can be represented as a causal theory $<C, P, E, \Sigma>$ where C corresponds to propositions concerning state variable values and so represents the possible causes; P corresponds to the set of possible port variable values; E corresponds to possible effects (i.e. observable values - here observable port values as $E \subseteq P$.); and $\Sigma$ to the behavioural model relating the causes to the effects (i.e. how state variable values and input port values determine output port values). Causes appear in explanations of symptom sets. Explanations are described later.

A typical elements of C is:

power_supply_1.working_status = not_working i.e. the "working_status" state variable of functional entity power_supply_1 (or synonymously unit power_supply_1 ) has value not_working.

Following others (e.g. Dousson 92) we will use a propositional reified logic to help express our models. Let $D \approx C \cup P$ be the set of propositions which are temporally qualified. D for example includes literals of the form F.O=abnormal, i.e. output port O of unit F has value "abnormal", and F.self=not_working for the self state variable of unit F. Elements of D can be qualified by predicates TRUE, FALSE, ON, and OFF.

For any proposition X of D and for any time t, either TRUE(X, t) or FALSE(X, t) holds. Additionally TRUE(X, t, t') means that X is TRUE from t till t', i.e. TRUE(X , t, t') $\equiv \forall \tau \in (t, t']$ , $TRUE(X, \tau)$. We will sometimes need different time points to be included and excluded and so define:

TRUE(X, t, t')$\equiv$TRUE]](X, t, t') $\equiv \forall \tau \in (t, t']$ , $TRUE(X, \tau)$
TRUE[](X, t, t') $\equiv \forall \tau \in [t, t']$ , $TRUE(X, \tau)$ etc.

In the process of generating explanations each element of C may be augmented with a temporal constraint. An example of such a statement is:

TRUE]](F. self= working, -∞, 20)

meaning that F is working during the interval (-∞, 20].

We need to represent the event of a state variable going to a not-working or to a specific fault mode state. Once a *state* variable has changed to not-working, or to a more specific fault mode, it is assumed to remain there. In what follows state variables are usually thought of as having domain { working, not_working}.

Events occur at an instant of time, and are expressed by the predicate ON. ON(X, t) $\equiv$
$\exists t, t' : (t < t < t') \wedge FALSE(X, t, t)$
$\wedge TRUE(X, t, t')$

An element of D can now appear in expressions expressing transitions. For an element F.self=not_working $\in$ C this could be of the form
$\exists t \in [t_1, t_2]$: ON(F.self= not_working, t)

## 2.1  SOME ASSUMPTIONS

Behaviour may result in temporal delays and this complicates the issues as there is uncertainty associated with the times till an effect occurs as well as the uncertainty associated with whether the event occurs at all. To allow a simple understanding of the uncertainty aspects the two aspects of uncertainty are decoupled as far as possible. The uncertainty associated with the antecedent determines if the rule is applicable but does not express any beliefs about when the effects will occur. The latter is expressed through the delays associated with the rules. In general the antecedent needs to contain predicates about temporal relationships but this will be minimised.

The approach will be described in the context of binary domains where components can be either working or not-working and ports can be either normal or abnormal. So whilst there may be a rich underlying functional model involving, say differential equations and control actions, we have mapped this model into a simpler functional model where the inputs and outputs refer to normal and abnormal. The system is assumed initially to be in a fully working state with all inputs and outputs normal. This does not mean that the system is static. Symptoms are also preclassified in terms of normal or abnormal. The richer model is some form of simulation and is used to map inputs and outputs in the simpler model to normal or abnormal by reasoning forwards from the inputs (if there are any) and hypothesized correct working behaviour in the richer model. This mapping of a richer functional model into a simpler one is not necessarily easy but common as it generally not possible to perform calculations regarding beliefs, particularly abduction, on the richer models.

We assume that a cause (or more precisely a state variable) may change only once - from working to not-working or to some specific fault mode. Different causal paths may lead to an output port which may be abnormal or normal. The effect of multiple reinforcements of "abnormal", say, through the different paths at possibly different times can have an effect on any rules which have this port in its antecedent.

When events are considered the process can become very complex even on such simple models. To simplify matters further some assumptions about the world will be made. The only abnormal symptoms at a port we will analyse is the first transition to abnormal. We will not be analysing signals of symptoms which go from normal to abnormal back to normal etc. This is quite reasonable given the abstraction level we are working at. Furthermore if the input is abnormal we have no way of saying what is normal output. We will look at the modeling of an uncertain link and a "noisy or" model depending on this assumption. Despite the simplicity of the model it is relevant to many applications.

A state variable X, say, will have domain:

$\{\exists_t:$ ON(X= not_working, t), TRUE][(X=working , -∞, +∞) \}

or if X is a port then the domain will be

$\{\exists_t:$ ON(X= abnormal, t), TRUE][(X=normal , -∞, +∞) \}

The time of the last symptom of interest can be substituted for +∞ if wanted.

We will usually write {x, ~x } for brevity, or {working, not_working}, or {normal, abnormal} but they have the interpretation above. Notice that the domain does not contain any information about *when* X becomes not_working or abnormal. Similarly the rules which give the semantics of the uncertainty rules do not involve quantification of time delays. This does not however mean that we do not use given information on time delays, just that the uncertainty semantics does not depend on knowing about the times.

In order to keep the modeling simple, temporal relationships are not used to specify the degree of coincidence of events, rather lower and upper bounds are placed on the delay. The approach could be extended naturally to include statements about temporal coincidence. A consequence of the restriction is that whilst temporal synergy is not ignored (positive synergy may for example allow a reduced upper bound to delays) it is not exploited as fully as it could. Importantly however it is not missed in explanations even though a more refined approach, naturally, would have more discrimination.

## 2.2 A SINGLE LINK

Take the case where an input port X affects an output port Z. Here we are looking at the case often modeled by a single link in an Bayesian network by four conditional probabilities.

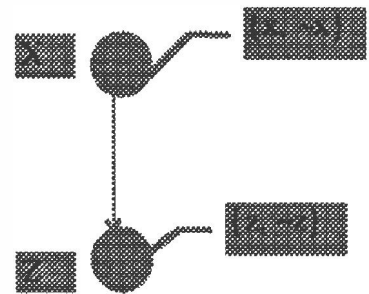

Figure 2: A link

In engineering systems the link is a functional entity and as such has behaviour. It is incompletely modeled. P[Z

=abnormall X=abnormal] is affected by the 'kind' of abnormality at X and the nature of propagation through the link, whilst P[Z=abnormall X=normal] essentially only refers to a "spontaneous" fault, ~y say, occurring in the link. If explanations involving the link functional entity are sought and if there are delays associated with this "fault" it can pay to model the link with an explicit "fault" state variable. (It is not being said that this cannot be modeled using a Bayesian link by adding an explicit cause. Rather it is arguing that the explicit cause often needs to be added and discussing the distribution with which to combine the effects from upstream and the explicit cause - when there are, and when there are not, temporal dependencies.)

A simple model which expresses the only ways that Z may become abnormal in terms of input port X and the state variable Y with domain $\{y, \sim y\}$ is

$$\sim x \wedge y \wedge \alpha \quad \vee \quad \sim y \leftrightarrow \sim z \qquad (A)$$

where $\sim x = \exists_t$: ON(X= abnormal, t), $\sim y = \exists_t$: ON(Y= not_working, t) and y = TRUE][(Y=working , $-\infty$, $+\infty$).

The model only involves one uncertain context assumption $\alpha$ (a 'causation event') which can be read as 'an abnormality at X is of the kind to produce an abnormality at Z when Y is normal' and is a reflection of the granularity of the model. Y is used to model effects on Z which are not caused by X. If Y were not present then P[Z=normallX=normal]=1. ~y may hold for different values of the input variable X; ~y covers the cases ~x $\wedge$ ~y and x $\wedge$ ~y. In fact the delay associated with ~y need not depend on the cases is may be taken as exactly 0 as Y is simply used to represent the functioning of the unit and Y has no parents . The delay associated with $\sim x \wedge y \wedge \alpha$ is specified by the user.

More generally we may have explicit delays associated with the terms in a rule (such as $\sim x \wedge y \wedge \alpha$ and $\sim y$) which can be used to compute intervals in which the events must have occurred. For example the following rule may actually be used

$\exists t^* \in [t + t_1, t + t_2]$:ON(X= abnormal, t) $\wedge$

TRUE][(Y=working , $-\infty$, $+\infty$) $\wedge \alpha \vee$

$\exists t^* \in [t + t_3, t + t_4]$:ON(Y= not-working, t) $\rightarrow$ ON (Z=abnormal, t*)

though the uncertainty semantics comes from the simpler rule. In such a case the delay associated with ~y needs to be conservative to cover the cases included.

When reasoning from observations in order to hypothesise causes we will want to determine the causes of ~z at time t. Using the relationship $\sim x \wedge y \wedge \alpha \vee \sim y \leftrightarrow \sim z$ allows us to associate an interval with ~x (and $\alpha$) of [t - upper bound to delay of $\sim x \wedge y \wedge \alpha$, t-lower bound to delay of $\sim x \wedge y \wedge \alpha$] which is interpreted as "at some time within", i.e. $\exists$t such that at some time in the interval the literal holds; and an interval [0, t) is associated with y and is read as "holds during" " i.e. $\forall$t in the interval the literal holds; and an interval [t-upper bound to delay of ~y, t] with ~y ] which is interpreted as "at some time within".

The relationship complementary to (A) is:

$$\sim \alpha \wedge \sim x \wedge y \quad \vee \quad y \wedge x \quad \leftrightarrow \quad z \qquad (B)$$

This states the conditions under which Z remains normal. (*Not* how Z becomes normal.) The relationship (B) is necessary for reasoning backwards from z. $\sim \alpha \wedge \sim x$ could be read "there is a case (or indeed some cases) when the input is abnormal, but for all of these cases they were of the wrong kind" to cause abnormality.

In such a complementary rule the concept of a delay has no meaning. *Only* rules which predict abnormal port values have user specified delays. However when reasoning from an observation z (i.e. z is normal and has always been normal) we can still associate intervals with the literals. If the literal is positive then this corresponds to a state variable or port value always being normal and so are given temporal intervals [0, t ) (or ($-\infty$, t ) if we imagine the system starting not at time 0, but at some period in the past before any symptoms occur). These intervals are now read as "during" i.e. $\forall$t in the interval the literal holds. If the literal is negative then the same interval is given (namely [0, t ) ) but the interpretation is "at some time", i.e. $\exists$t such that at some time in the interval the literal holds. The different interpretations of the intervals are used when intersections and unions of the intervals are taken.

We wanted a simple model of a link as we want to reason backwards without generating unnecessary possibilities in order to compute possible temporal intervals to associate with fault hypotheses. The model (A) and (B) is relevant in many cases. A rather informal justification is given below. The full conditional probability distribution would be:

$\sim x \wedge y \wedge \alpha \quad \vee (\sim x \wedge \sim y) \wedge \beta \vee x \wedge y \wedge \gamma \vee x \wedge \sim y \wedge \delta \leftrightarrow \sim z$

$x \wedge y \wedge \gamma$ is not worth having as we are happy to believe that in the case $x \wedge y$ then z must always hold. The most problematic term is $\sim x \wedge \sim y \wedge \beta$. Whilst the term $\sim x \wedge \sim y \wedge \beta$ can lead to a multiple fault explanations for $\sim z$ (and we *are* interested in multiple fault explanations) the difference between this multiple fault and others covered later is that Y is a state variable and so has no upstream causes. Computationally it is easy to create the explanations for the joint occurrence of $\sim y$ and the causes of $\sim x$ *simply by replacing* $y \wedge \alpha$ *by* $\sim y$ *in all explanations* generated which contain y (and $\alpha$). The belief for an explanation containing causes of $\sim x$ and $\sim y$ must be less than the belief of the explanation containing causes for $\sim x$ and y so long as $P[y \wedge \alpha] > P[\sim y]$ This is very reasonable and so will apply in most models. So such explanations already generated should be kept until their belief, when $\sim y$ replaces $y \wedge \alpha$ , is greater than any others. Hopefully we would have identified the causes before this stage. Similarly for the case $(\sim x \wedge \sim y) \wedge \sim \beta$ causing z. The temporal intervals to associate with the causes in the explanations need to be revised but this is not discussed here.

## 2.3 TWO INPUT PORTS AFFECTING AN OUTPUT PORT

Here there are two inputs with domains {normal, abnormal}. We will assume that uncertainty arises through granularity rather than through failure of the functional unit. A functional model with multiple inputs and with state could be built from links as previously described and the unit to be described.

Under the assumption of no state if both inputs are effectively normal then the output must be normal. The most simple model (under our assumptions) is:

$$x \wedge y \;\; \vee \;\; (\sim x \wedge \sim \alpha) \wedge y \;\; \vee \;\; x \wedge (\sim y \wedge \sim \beta) \;\; \vee \;\; (\sim x \wedge \sim \alpha) \wedge (\sim y \wedge \sim \beta) \leftrightarrow z$$

with complementary relationship

$$\sim x \wedge \alpha \;\; \vee \sim y \wedge \beta \leftrightarrow \sim z$$

When reasoning backwards from z the interval associated with $\sim x$ or $\sim y$ is [0, t) and interpreted as at some time within. The intervals created for y and x are interpreted as during.

Now we have to make some assumptions about the nature of the propagation. We will assume that two events can occur at Z (i.e. the effect is not masked) If the port Z fans out to affect many downstream functional entities then the two causes of $\sim z$ may cause symptoms at different times. However if Z is an observable then the second event is *masked as far as observation* is concerned but not as regards further propagation. So abduction from an *observed* symptom is finding the explanations of the first 'symptom'. Predictions of later abnormalities at the point where an abnormal symptom is observed are not necessarily inconsistent.

If there are some time delays, and even in the case where all the delays are zero, it is possible that a more liberal notion of the "right kind of $\sim x$" and the "right kind of $\sim y$" is appropriate when $\sim x$ and $\sim y$ hold simultaneously as there may be some synergy which allows a broader (reduced is conceptually possible, but ignored here) context to trigger the effect. This is called context synergy.

A richer model which allows for a broader context to be applicable when abnormality is present at both inputs is:

$$(\sim x \wedge \alpha \wedge y) \vee (\sim x \wedge \alpha) \wedge (\sim y \wedge \sim(\beta \vee \sigma)) \quad \vee \quad (x \wedge \sim y \wedge \beta) \vee (\sim x \wedge \sim (\alpha \vee \psi)) \wedge (\sim y \wedge \beta) \quad \vee \quad (\sim x \wedge (\alpha \vee \psi)) \wedge (\sim y \wedge (\beta \vee \sigma) \leftrightarrow \sim z \quad (1)$$

or equivalently

$$(\sim x \wedge \alpha) \vee (\sim y \wedge \beta) \;\vee\; (\sim x \wedge \psi) \wedge (\sim y \wedge \sigma) \leftrightarrow \sim z \;\; (1a)$$

The last term gives the case where the joint occurrence of $\sim x$ and $\sim y$ are necessary for $\sim z$. This reduces the to previous model if $\psi$ and $\sigma$ do not hold. We may associate with $\psi$ and $\sigma$ a probability (say) of $P[\sigma]$ and $P[\psi]$.

(1) has complement:

$$x \wedge y \vee (\sim x \wedge \sim \alpha) \wedge y \; \vee \; (\sim y \wedge \sim \beta) \wedge x \; \vee \quad (\sim x \wedge \sim(\alpha \vee \psi)) \wedge (\sim y \wedge \sim \beta) \quad \vee \quad (\sim y \wedge \sim(\beta \vee \sigma)) \wedge (\sim x \wedge \sim \alpha) \leftrightarrow z \quad (2)$$

These relationships say *nothing* about the times and delays. It just specifies all the contexts that can cause $\sim z$. Here $\psi$ and $\sigma$, where $\psi \wedge \alpha = \perp$ and $\sigma \wedge \beta = \perp$, are the additional contexts where $\sim x$ can cause $\sim z$ etc. (Other models are possible.)

Even if both $\sim x$ and $\sim y$ occur they do not need to overlap in the times of their effects, though there must come a time when they are both true, because of our assumption that they cannot return to either x or y. However synergy may also make effects quicker or slower. This is temporal synergy. The time to the effect is also influenced by the degree of temporal coincidence of input values. For example in (1) if ~x occurs long before ~y and the delay associated with ~x is not long, then the delay will be that of ~x even if there would be temporal synergy when both ~x and ~y hold. The delay associated with a term such as ~x ∧ (α ∨ ψ) ∧ ~y ∧ (β ∨ σ)    depends on how close ~x and ~y happen.

An advantage of the representation of (1a) is that the time difference of ~x and ~y need not be expressed explicitly. From (1) it can be seen that minimum delay associated with each of the first two terms of (1a), i.e. (~x ∧ α) and (~y ∧ β), must be equal to zero. The upper bounds to these two terms will come from the bounds associated with (~x ∧ α) ∧ y and (~y ∧ β) ∧ x respectively. The last term of (1a) only has a meaning when ~x and ~y temporally overlap and needs to be specified by the modeler. The effect can only start when both ~x and ~y hold so it may be non zero.

If there is temporal synergy we could, as well as dividing into terms based on context (as has been done), divide into terms based on the time between ~x and ~y. Then more precise intervals could possibly be attributed to the terms. However the boundaries between different cases are not easy to specify, and in the interests of parsimony we do not use temporal conditions in the terms. This means that the intervals in the relationships must specify (in the positive synergy case) the lower bounds based on maximal temporal coincidence and upper bounds on minimal temporal coincidence. So unless temporal conditions are used as part of the terms any temporal synergy is partially masked by the different degrees of coincidence. The approach tries to keep the models the user has to specify simple.

## 3 FINDING EXPLANATIONS FROM SYMPTOMS - IMPLEMENTATION

The reasoning process produces abductive *explanations* for a set of observations $O \subseteq E$. An abductive explanation for a single observation is simply a special case which is important in the implementation. An explanation is denoted by $\varepsilon$ and is a conjunction of literals from C, each augmented by a temporal constraint and uncertainty value, such that each $\varepsilon$ is consistent with $\Sigma$ . ($\Sigma$ includes all the relationships between input ports, state variables and output ports and also the temporal constraints. )
$\Sigma \cup \varepsilon \vdash O$ to some degree of belief > 0.

An example of an explanation is:
TRUE(unit5.working-status = working, -∞, 10) ∧
$\exists t \in (-\infty, 20]$ :ON(unit1.working-status = not-working, t) probability 0.05

The $\varepsilon$ generated are not mutually exclusive (c.f. Poole).This allows explanations to retain their interpretation simplicity.

The condition $\Sigma \cup \varepsilon \vdash O$ to some degree of belief is satisfied by the construction process described since the model is assumed covering A *cautious explanation set* x, for O is a disjunction of all the abductive explanations for O i.e. $\xi = \vee \varepsilon_i$ where each $\varepsilon_i$ is an abductive explanation. An option available in small applications is to produce a cautious explanation set, though for non trivial problems only a subset can be practically generated. In the current implementation a cost bounded ATMS    (CBATMS) (Ngair and Provan, Bigham, De Kleer) is used as a focusing mechanism to select low cost explanations first. A polynomial time behaviour can be achieved using a cost function and assuming the ATMS network contains no cycles and has $O(q^2)$ connections for q network nodes.

From the definition of an explanation it can be seen that the multiple fault model is being considered. In the implementation the single fault model is included efficiently as a special case because of the incremental approach.

If we have observations $o_1, o_2, .. o_n$, and corresponding cautious explanation sets for each individual observation $\xi_1, \xi_2, \xi_3 ,...... \xi_n$ . Since all observations are true and the rules are covering then

$$\vdash o_1 \wedge o_2 \wedge ...... \wedge o_n \equiv \xi_1 \wedge \xi_2 \wedge ....... \wedge \xi_n$$

Temporal intersections are used to eliminate explanations which are impossible because of the timings of the symptoms.

## 3.1  PROBABILITY

An efficient algorithm based on an extension of Ngair and Provan's cost bounded ATMS to incorporate elimination of inconsistent temporal intervals can be used to construct explanations. This can be used even in cases where there are complementary assumptions, such as a and ~a, which is common in uncertainty calculations for useful cost functions based on the number of not-working assumptions and on possibilistic logic. (See Bigham 95) The basic consideration is that the cost function must be such that resolution of environments cannot create a lower cost environment.

However probability theory (in a general context) does not satsify the conditions on a cost function to ensure that environments are computed in a manner which yields an incremental algorithm and efficient blocking and unblocking of propagation in the ATMS.

However if we include all causation events as part of C, the set of causes, then it is not difficult to show that monotonicity is assured in generation of the explanations in the CBATMS. (We still cannot replace causation events with numbers as could be done in the possibilistic case as it is important not to multiply repeatedly when environments are combined. In the possibilsitic case this is not a problem as minimums and maximums are taken.)

The only problem is whether including causation events yields an adequate explanation. In general it does not. It can be argued that in practice when the timing of events is considered, it often is adequate. If we consider the symptom of payload shedding given earlier, there are two ways that this can happen (via the electrical path and via the thermal path). An explanation in terms of just causes (e.g. Overvoltage Transistor Failure) is implicitly a disjunction consisting of the different ways payload shedding could happen given the same root causes(s). Each disjunct would contain the same causes but different causation events corresponding to different paths. However the different causal paths make payload shedding occur at different times and the symptom has been triggered by events along *one* of the paths. The events along the other path may not even have happened yet. An explanation should be in terms of causes and causation events from a single causal path. So combination is not necessary, and indeed not desirable.

## 4  CONCLUSION

A knowledge representation language that can be used to represent diagnostic problems that involve uncertainty, and imprecise propagation delays has been presented.

The implementation approach to uncertain reasoning uses an incremental cost bounded algorithm where appropriate cost functions can be used to limit the computation in the context of the application requirements. A correlation system has been implemented in C++.

### Acknowledgments

Part of this work was carried out in the DRUMS II and UNITE projects.